\documentclass[11pt]{article}
\usepackage{amsmath}
\usepackage{multirow}
\usepackage{hyperref}
\newcommand{\qms}[1]{\textcolor{blue}{#1}}

\usepackage[final]{acl}

\usepackage{times}
\usepackage{latexsym}

\usepackage[T1]{fontenc}

\usepackage[utf8]{inputenc}

\usepackage{microtype}

\usepackage{inconsolata}

\usepackage{graphicx}

\title{From Answers to Questions: EQGBench for Evaluating LLMs' Educational Question Generation}

\author{
 \textbf{Chengliang Zhou\textsuperscript{1}},
 \textbf{Mei Wang\textsuperscript{1}},
 \textbf{Ting Zhang\textsuperscript{1}},
 \textbf{Qiannan Zhu\textsuperscript{1}},
 \textbf{Jian Li\textsuperscript{1}},
 \textbf{Hua Huang\textsuperscript{1}},
 \\
 \textsuperscript{1}School of Artificial Intelligence, Beijing Normal University
}

\begin{document}
\maketitle
\begin{abstract}
Large Language Models (LLMs) have demonstrated remarkable capabilities in mathematical problem-solving. However, the transition from providing answers to generating high-quality educational questions presents significant challenges that remain underexplored. To advance Educational Question Generation (EQG) and facilitate LLMs in generating pedagogically valuable and educationally effective questions, we introduce EQGBench, a comprehensive benchmark specifically designed for evaluating LLMs' performance in Chinese EQG. EQGBench establishes a five-dimensional evaluation framework supported by a dataset of 900 evaluation samples spanning three fundamental middle school disciplines: mathematics, physics, and chemistry. The dataset incorporates user queries with varying knowledge points, difficulty gradients, and question type specifications to simulate realistic educational scenarios. Through systematic evaluation of 46 mainstream large models, we reveal significant room for development in generating questions that reflect educational value and foster students' comprehensive abilities.
\end{abstract}

\section{Introduction}

From the advent of GPT-3 to the latest breakthroughs with ChatGPT and GPT-4, Large Language Models (LLMs) have demonstrated extraordinary capabilities in understanding complex queries and generating human-like responses, particularly in the realm of mathematical problem-solving. However, a fundamental shift emerges when we move from answers to questions: can these powerful models, adept at providing solutions, master the more challenging task of question generation within educational contexts?

In the educational domain, the ability to generate high-quality questions is a cornerstone of effective pedagogy and learning. While various Automatic Question Generation (AQG) methodologies \cite{aqg1,aqg2,aqg3} have emerged in recent years, existing approaches predominantly focus on deriving questions from predetermined answers and their corresponding contextual information, rather than addressing the distinctive requirements inherent in educational question generation (EQG). This distinction is of critical importance: EQG emphasizes the generation of questions based on specific pedagogical requirements and learning objectives. Furthermore, EQG must extend beyond superficial factual recall to cultivate higher-order cognitive competencies, encompassing conceptual understanding, reasoning capabilities, problem-solving proficiencies, thereby establishing more rigorous requirements for question generation systems.

\begin{figure}[t]
  \includegraphics[width=\columnwidth]{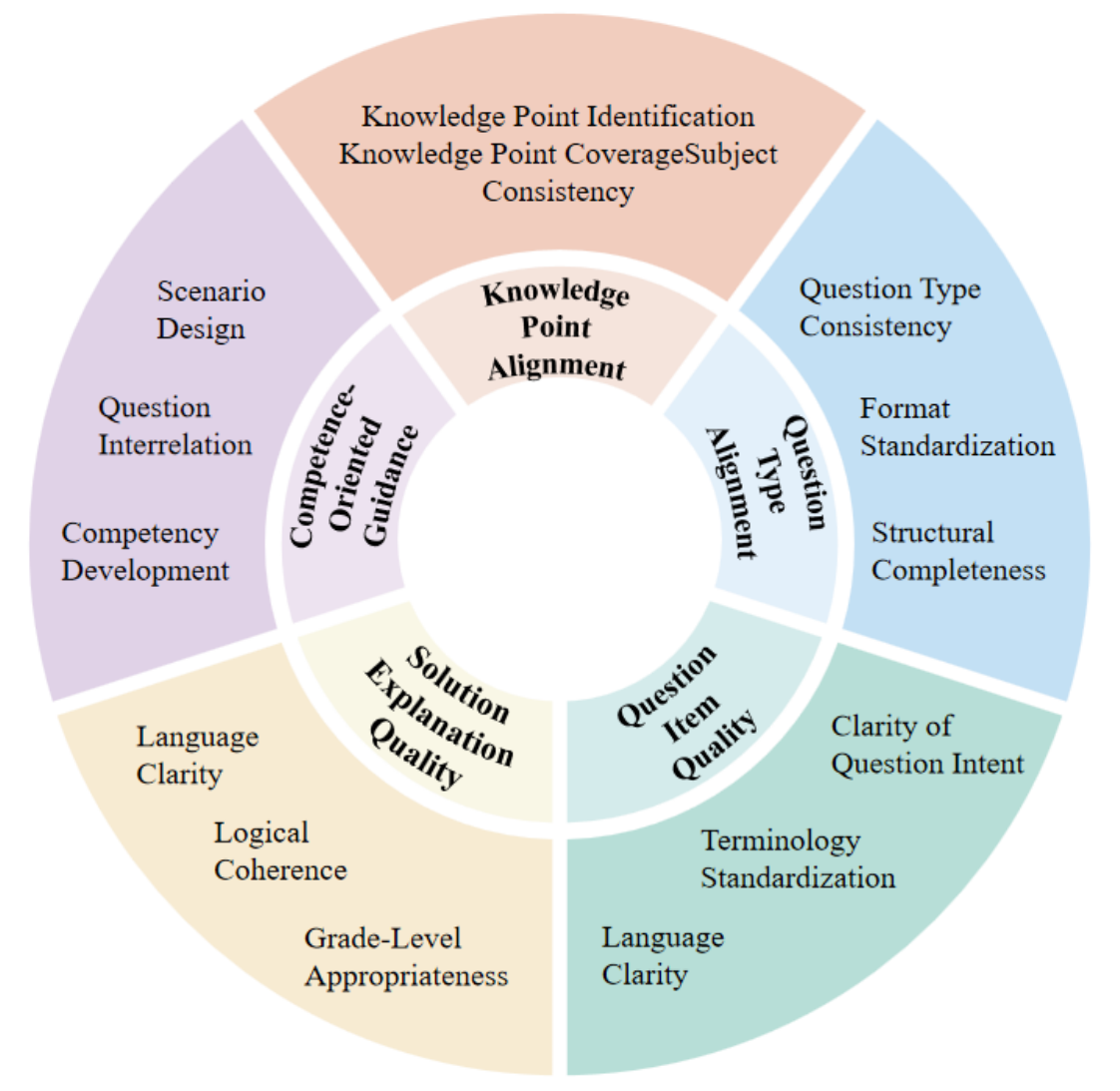}
  \caption{The design of EGQBench’s evaluation dimensions along with their corresponding detailed metrics.}
  \label{fig1}
\end{figure}

To advance the rapid development of EQG, comprehensive evaluation benchmarks are crucial. However, many existing benchmarks for Automatic Question Generation (AQG) rely on n-gram-based metrics like BLEU and ROUGE. This evaluation paradigm is fundamentally misaligned with the objectives of education. Such metrics reward surface-level lexical similarity, but the value of an educational question lies not in its phrasing, but in the cognitive processes it elicits. For instance, these metrics cannot distinguish between a simple fact-recall question and a complex problem that requires multi-step reasoning, application of concepts, or higher-order thinking. An effective educational question guides a student through a specific problem-solving pathway, a dimension entirely invisible to text-similarity algorithms. Consequently, relying on these metrics hinders progress towards generating questions that are genuinely effective for teaching and learning. 

To address this critical gap, we introduce EQGBench, a comprehensive benchmark specifically designed to evaluate the capability of models to generate high-quality educational questions in Chinese. EQGBench is supported by a carefully curated dataset of 900 evaluation samples spanning three fundamental middle school disciplines: mathematics, physics, and chemistry, with 300 samples evenly distributed across each subject. The dataset incorporates diverse user queries with varying knowledge points, difficulty gradients, and question type specifications to authentically simulate real-world educational scenarios. Furthermore, it provides a multi-dimensional evaluation framework deeply aligned with educational objectives, encompassing knowledge point alignment, question type alignment, question item quality, solution explanation quality, and a key dimension of competence-oriented guidance. 
By translating question quality into a series of quantifiable and interpretable evaluation dimensions, we offer a fine-grained analysis of the strengths and limitations of models as they transition from being "answer providers" to "question creators."

Using EQGBench, we conducted a comprehensive evaluation of 46 mainstream LLMs, including models from the ChatGPT, DeepSeek, and GLM series. Our experimental results reveal that models exhibit minimal inter-disciplinary variance on fundamental comprehension tasks, while models with larger parameter counts hold a distinct advantage in tasks demanding higher-order logical reasoning. A key finding is that competence-oriented guidance represents a significant weakness across the board. 

The main contributions of this paper are as follows:

\begin{enumerate}
    \item We construct a high-quality EQG dataset that simulates real-world scenarios, covering three core subjects in middle school: mathematics, physics, and chemistry.
    \item We design a five-dimensional evaluation framework to comprehensively measure both the content quality and the pedagogical value of the generated questions.
    \item We perform a systematic evaluation of 46 mainstream LLMs, including ChatGPT, DeepSeek, and GLM. Through human study, we validate the scientific rigor and practical utility of EQGBench for assessing question generation.
\end{enumerate}
\section{Related Work}

\subsection{Question Generation}
Early research in question generation (QG) primarily relied on template-based methods and neural sequence-to-sequence (Seq2Seq) models. Template-based approaches populate predefined sentence structures with knowledge points, but they suffer from poor flexibility and produce monotonous content \cite{ali2010automatic,mitkov2003computer,heilman2010good,mostow2009generating}. While neural Seq2Seq models were capable of generating relevant questions from a given context, they demonstrated limited ability in terms of creativity and understanding complex instructions \cite{10.1007/978-3-319-73618-1_56,zhao2018paragraph,10.5555/3454287.3455457,10.1145/3397271.3401037}. Leveraging their powerful zero-shot and few-shot capabilities, LLMs can now generate questions tailored to user needs through well-designed prompts, offering high flexibility and versatility \cite{MAITY2025100370,maity2024futurelearningagegenerative}.

Despite these significant technological advancements, the evaluation of LLM-based question generation has lagged. A number of Chinese LLM benchmarks have recently emerged, such as the GAOKAO Benchmark \cite{zhang2024evaluatingperformancelargelanguage}, which uses national college entrance exam questions to assess problem-solving skills; C-EVAL \cite{NEURIPS2023_c6ec1844}, a comprehensive Chinese language evaluation suite; CMMLU \cite{li2024cmmlumeasuringmassivemultitask}, a multi-disciplinary benchmark; and FinEval \cite{guo2024finevalchinesefinancialdomain}, an assessment of financial knowledge. However, a common thread in these works is their focus on evaluating models' knowledge reserves and problem-solving abilities. They fall short of assessing a model's capacity to generate creative and insightful questions that align with curriculum standards under specific pedagogical objectives.

\subsection{Question Evaluation}

Question evaluation is primarily divided into manual and automated evaluation. Manual evaluation requires experts, such as teachers with specialized knowledge, to provide comprehensive scores across multiple dimensions. For instance, some studies have utilized crowd-workers to score questions on a 1-to-5 scale \cite{du-cardie-2017-identifying,du2018harvestingparagraphlevelquestionanswerpairs}. Similarly, MATHWELL \cite{christ2024mathwellgeneratingeducationalmath} is a recently proposed framework that guides manual annotation. However, manual evaluation is costly, time-consuming, difficult to scale, and its results can be influenced by evaluator subjectivity, rendering it unsuitable for the rapid evaluation of numerous models. Traditional automated metrics like BLEU \cite{BLEU}, ROUGE \cite{lin-2004-rouge}, and METEOR \cite{banerjee-lavie-2005-meteor} score questions by calculating the n-gram overlap between generated and reference texts. These metrics primarily measure surface-level textual similarity and cannot effectively evaluate a question's logical coherence, solvability, or educational value. Consequently, their utility is extremely limited for tasks like question generation that demand high semantic and logical accuracy.

The paradigm of using LLMs for automated evaluation has emerged as a new research hotspot. Zheng \cite{10.5555/3666122.3668142} and Chiang \cite{vicuna2023}demonstrated the feasibility and reliability of using LLMs as judges. This approach has been applied to essay scoring \cite{kim2024languagemodelsevaluatehuman} and mathematics answer evaluation \cite{jiang2025llmsmathematicalreasoningmistakes,systems11070353}. Wang \cite{wang2024automaticanswerabilityevaluationquestion} proposed the PMAN metric, which prompts an LLM to answer a question it has generated to determine the question's validity. In the QG domain, some studies have also begun to explore deeper evaluation dimensions. EduBench \cite{xu2025edubenchcomprehensivebenchmarkingdataset} evaluates models within a broader educational context, while Dr.Academy \cite{chen2024dracademybenchmarkevaluatingquestioning} assesses question generation capabilities based on Bloom's Taxonomy.

Although these studies represent positive progress, they often fail to connect with practical, real-world requirements. In contrast, EQGBench is a comprehensive question generation benchmark directly linked to the core pedagogical principles and curriculum requirements of middle school education. This direct alignment ensures that our evaluation results carry greater practical significance and instructional relevance.

\section{Dataset Construction}

Educational question generation in real-world educational settings is a highly complex and contextualized task. Its demands extend far beyond simple knowledge point retrieval, posing a severe challenge to the text understanding and generation capabilities of existing automated systems. There is currently a lack of systematic evaluation benchmarks specifically designed for these complex educational needs. 

To bridge this gap, we introduce EQGBench, a comprehensive evaluation dataset designed to systematically assess LLMs' educational question generation capabilities. EQGBench comprises 900 high-quality evaluation samples evenly distributed across mathematics, physics, and chemistry. Through structured template design and dynamic information filling, the dataset generates diverse user queries with varying knowledge points, difficulty gradients, and question type specifications across multiple educational contexts—including teacher lesson preparation, personalized student practice, and parental guidance.


\begin{table*}[ht]
  \centering
  \footnotesize
  \renewcommand{\arraystretch}{1.4}
  \begin{tabular}{|c|p{0.4\textwidth}|p{0.4\textwidth}|}
    \hline
        ~ & Template Sample & Specific Sample  \\ \hline
       \multirow{1}{*}{Sample 1} & I am self-studying \qms{\{grade\}} \   \qms{\{subject\}}, and I have currently reached the \qms{\{knowledge\}} section. I would like a self-assessment exercise in the form of a \qms{\{question\_type\}}. & I am self-studying middle school mathematics, and I have currently reached the basics of rational numbers section. I would like a self-assessment exercise in the form of a solve-and-explain question.  \\ 
        \hline
        Sample 2 & I am a \qms{\{subject\}} student teacher, and I need to design interactive board work for tomorrow's demo lesson. Please include a \qms{\{question\_type\}} in the \qms{\{knowledge\}} section, with a difficulty level of \qms{\{difficulty\}}, in accordance with the \qms{\{grade\}} curriculum. & I am a mathematics student teacher, and I need to design interactive board work for tomorrow's demo lesson. Please include a single-choice in the maximum value problem of $y = ax^2 + bx + c$ section, with a difficulty level of easy, in accordance with the middle school curriculum.  \\
        \hline
        Sample 3 & My child is in \qms{\{grade\}} this year, and he tells me that he can never understand the \qms{\{knowledge\}} section in \qms{\{subject\}}. Could you provide \qms{\{num\}} \qms{\{question\_type\}} questions for practice? & My child is in middle school this year, and he tells me that he can never understand the simplifying absolute values within a range section in mathematics. Could you provide 2 fill-in-the-blank questions for practice?  \\ \hline
        Sample 4 & I need to consolidate the \qms{\{knowledge\}} section in \qms{\{grade\}} \qms{\{subject\}}. Can you give me \qms{\{num\}} questions with \qms{\{difficulty\}} level? & I need to consolidate the real numbers section in middle school mathematics. Can you give me 3 questions with medium level?  \\ \hline
  \end{tabular}
  \caption{Example data from EQGbench . Each template type embodies a user scenario and their specific needs, posing targeted requests from the perspective of a teacher, student, or parent, alongside generic queries without a specific persona. }
  \label{example data}
\end{table*}

\subsection{Template Construction}

To ensure that EQGBench's templates comprehensively cover the demands of real-world teaching scenarios while guaranteeing the quality and diversity of the instructions, we first invited several veteran middle school teachers to design approximately 40 initial instructions for EQG across the core subjects. 
Based on these 40 human-designed instructions, we employed a three-step process of parameterization, rewriting, and analogical generation to create a larger and more linguistically diverse prompt set.

\paragraph{Parameterization} We deconstructed the initial instructions, abstracting core requirements such as academic stage, subject, knowledge point, question type, difficulty level, and the desired number of questions into parametric variables. This process formed a set of structured base templates.

\paragraph{Stylistic Rewriting} We utilized various LLMs, including Doubao, Qwen, and DeepSeek, to rewrite these base templates from multiple perspectives—such as those of a teacher, a student, and a parent—to introduce rich stylistic variations.

\paragraph{Analogical Generation} We used the rewritten templates as exemplars to prompt the LLMs to generate a larger corpus of new prompt templates through imitation.



\subsection{Template Filling and Data Generation}
To generate a diverse set of user prompts from the structured templates, we employed a stratified random sampling strategy. This method dynamically populates the multi-dimensional parameters within the templates—including academic stage, subject, number of questions, knowledge points, question type, and difficulty—based on predefined distributions. This process resulted in the final evaluation dataset of 900 instructions. Examples of the resulting data are shown in Table ~\ref{example data}.

The detailed information for the data filling is as follows:

\textbf{Grade:} Uniformly set to "middle school" to precisely match the pedagogical requirements of this compulsory education phase.

\textbf{Subject:} Covers three core science disciplines: "Mathematics," "Physics," and "Chemistry," with 300 samples per subject.

\textbf{Number of Questions:} Instructions request either a "single" question or "multiple" questions (specifically 2 or 3). For each subject, the ratio of instructions for single versus multiple questions is 260:40.

\textbf{Knowledge: }The knowledge points are derived from the official middle school curriculum for each subject. This knowledge base is organized into a hierarchical, tree-like structure where concepts are progressively detailed across logical tiers. Specifically, the mathematics knowledge system consists of 4 main levels leading to terminal knowledge points, while the physics and chemistry systems each have 5 levels.

\textbf{Question Type:} Three types of questions were specified: "Single-choice", "Fill-in-the-blank" and "Problem" distributed in a 4:3:3 ratio.

\textbf{Difficulty:} A differentiated distribution of difficulty was designed. Five types were included: "simple," "medium," "hard," "from easy to hard" (progressive), and "from hard to easy" (regressive), distributed in a balanced 1:1:1:1:1 ratio.

\subsection{Human Review}
To ensure the high quality of the final dataset, we performed a thorough manual review of all successfully generated instructions. The review process focused on three key aspects: smoothness of phrasing, accuracy of word choice, and format consistency. This step aimed to prevent issues such as content omissions, awkward phrasing, or formatting errors, ensuring that each prompt accurately represents a real user query scenario.
\section{Evaluation Metric Design}
As a critical application of educational technology, educational question generation poses a significant test for LLMs. The evaluation of this capability is complex because instructional content involves multi-layered knowledge systems, and different educational contexts have varying standards for question quality. Consequently, traditional evaluation methods struggle to comprehensively measure a model's question generation proficiency. Although recent studies have explored the feasibility of using LLMs for evaluation \cite{4.11,4.12,4.13}, a specialized evaluation framework for question generation capabilities remains underdeveloped. To address this, we have constructed a multi-dimensional, comprehensive evaluation framework to systematically measure the performance of LLMs on educational question generation tasks. This framework is built upon five key metrics: knowledge point alignment, question type alignment, question item quality, solution explanation quality, and competence-oriented guidance. Dimensions were rated on three levels—\textbf{Excellent}, \textbf{Good}, and \textbf{Poor}—corresponding to scores of \textbf{2}, \textbf{1}, and \textbf{0}, respectively.

\paragraph{Knowledge Point Alignment (KP)}
This dimension assesses whether the generated question accurately identifies and reflects the knowledge point(s) specified in the user's input, ensuring the question aligns with the designated topic.
\begin{itemize}
  \item \textbf{Excellent:} The generated question accurately and fully covers the user-specified knowledge point(s).
  \item \textbf{Good:} The question is correct in the broader knowledge area but does not align with the specific, detailed knowledge point requested by the user.
  \item \textbf{Poor:} The question completely fails to cover the specified knowledge point or pertains to a different academic subject.
\end{itemize}

\paragraph{Question Type Alignment (QT)}
This dimension evaluates whether the type of the generated question (e.g., choice, fill-in-the-blank, problem) matches the user's selection and adheres to the standard formatting requirements for the type. For example, a single-choice question should include four options; a fill-in-the-blank question should provide an underline, parentheses, or another clear indicator for the answer; a problem may be presented as a comprehensive problem that integrates various formats like selection or calculation.
\begin{itemize}
  \item \textbf{Excellent:} The question's type is identical to the user's specification and adheres to the standard format for that type.
  \item \textbf{Good:} The question's type is generally consistent with the user's specification, but there are minor errors in detail or formatting.
  \item \textbf{Poor:} The question's type is completely inconsistent with the user's specification, or the format is too disorganized to be identified.
\end{itemize}

\paragraph{Question Item Quality (QQ)}
This dimension assesses whether the generated question is expressed clearly, has an unambiguous objective, uses standardized terminology, and is solvable with a unique or definitive answer. This ensures that students can accurately understand the question's intent and complete the task.
\begin{itemize}
  \item \textbf{Excellent:} The question is clear, concise, and easy for students to understand.
  \item \textbf{Good:} The language of the question is ambiguous, or technical terms are used incorrectly.
  \item \textbf{Poor:} The language is confusing or unclear, with significant issues such as redundancy, logical fallacies, or typos.
\end{itemize}

\paragraph{Solution Explanation Quality (SQ)}
This dimension evaluates the correctness, rigor, and completeness of the explanation provided for the generated question. It also requires that the knowledge involved in the explanation is appropriate for the cognitive level and curriculum requirements of the target academic stage, and that the correct answer can be derived from the explanation.
\begin{itemize}
  \item \textbf{Excellent:} The explanation is correct, logically sound, and meets all requirements of the question.
  \item \textbf{Good:} The explanation contains logical leaps, lacks clarity, or has issues like repetition.
  \item \textbf{Poor:} The explanation process is flawed and cannot lead to the correct answer, or no explanation is provided at all (only the final answer is given).
\end{itemize}

\paragraph{Competence-Oriented Guidance (CG)}
This dimension evaluates whether the generated question integrates or simulates a realistic scenario, including but not limited to cultural contexts, practical subject applications, or real-life situations. It measures the question's value in guiding students to apply knowledge and develop higher-order competencies.
\begin{itemize}
  \item \textbf{Excellent:} The question incorporates a rich, contextual scenario that is directly relevant to solving the problem.
  \item \textbf{Poor:} The question is a purely abstract application of knowledge points, lacking any contextual design.
\end{itemize}

\section{Experiments}
\subsection{Experiment Settings}
For the responses of open-source models, we deployed models with smaller parameter sizes on a single NVIDIA A800 GPU server, each with 80GB of memory. The inference was carried out using the vLLM framework in a Python environment. For the closed-source models, we utilized a single NVIDIA 4060 GPU with 16GB memory and accessed various models via the API provided by the OpenAI library.

Uniform model parameters were set across all models for consistency. The temperature parameter was set to 0.6 to ensure higher randomness during question generation. The maximum output length was capped at 4096 tokens to prevent overthinking or excessive output. For models with a "thinking" mode, this feature was enabled to facilitate better question generation.

\subsection{Evaluation Details}
\begin{figure}[htb]
  \includegraphics[width=\columnwidth]{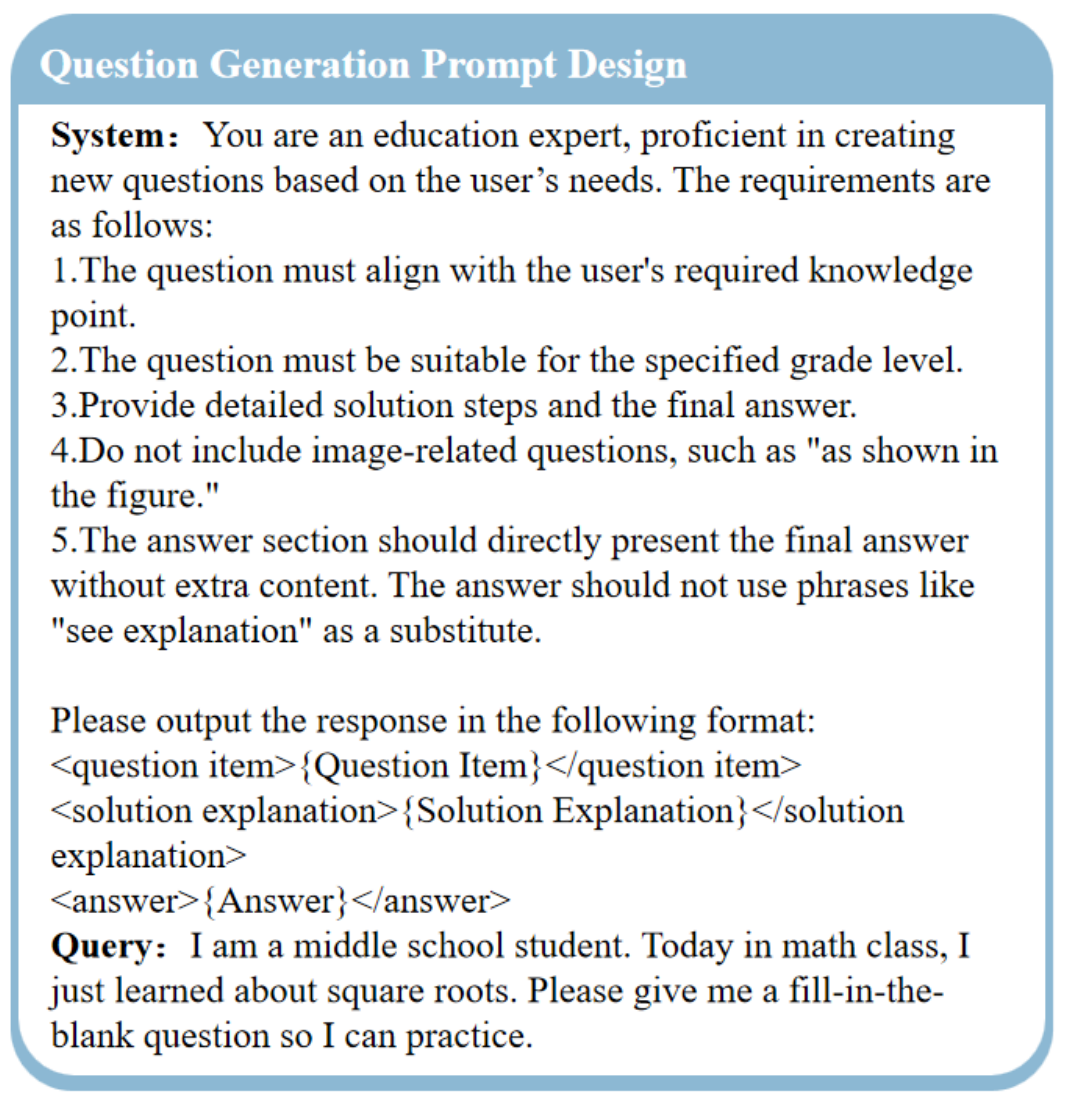}
  \caption{Representative prompt designs used in EGQBench to obtain model-generated questions tailored to user requirements.}
  \label{fig2}
\end{figure}

\paragraph{Response Generation}
To thoroughly examine the performance of LLMs of varying architectures and scales in intelligent question generation, we selected 46 mainstream models, including DeepSeek R1, ChatGPT, Qwen3, and Gemini, with parameter sizes ranging from 7 billion to hundreds of billions. These models include both specialized models focused on reasoning and general-purpose dialogue models, allowing us to comprehensively analyze the effects of model size, architecture, and capacity on question generation quality. The prompt used for response generation is shown in Figure ~\ref{fig2}.

\begin{figure}
  \centering
  \includegraphics[width=\columnwidth]{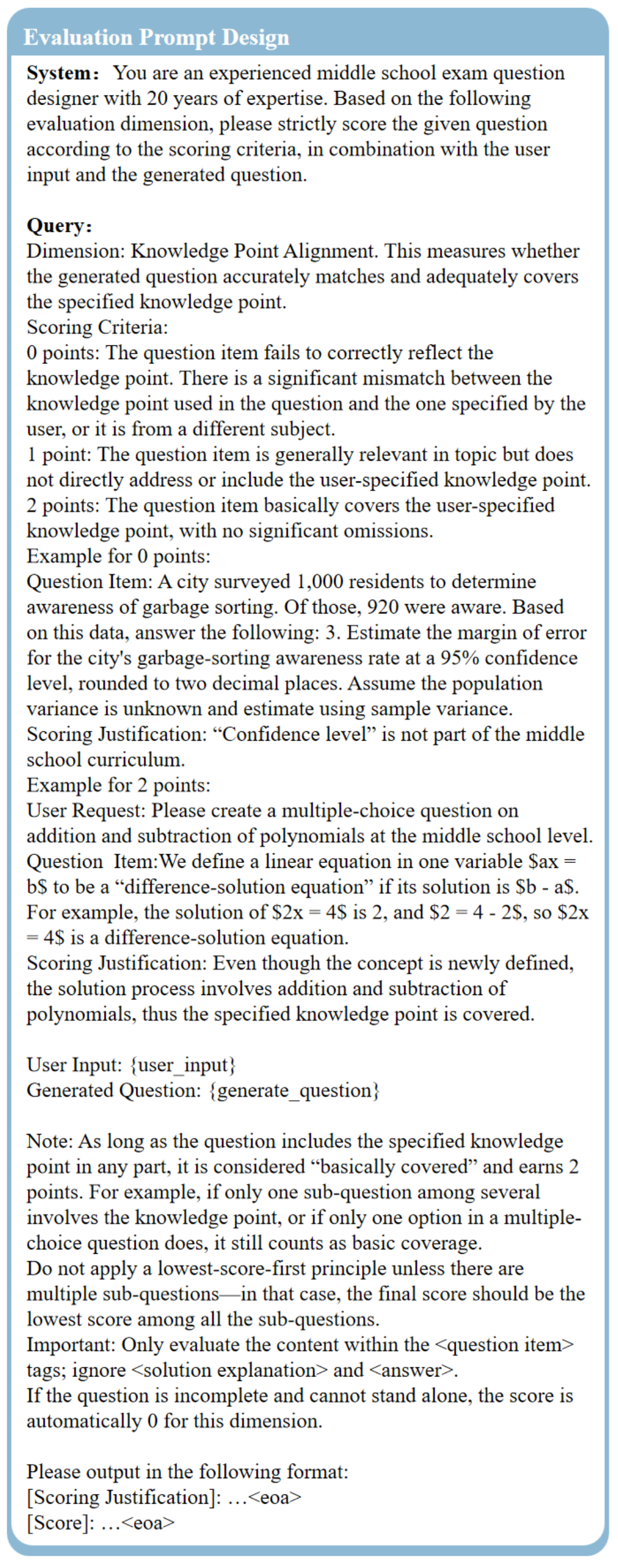}
  \caption{Representative prompt designs used in EGQBench to evaluate model-generated questions with respect to knowledge point alignment.}
  \label{fig3}
\end{figure}

\paragraph{Response Evaluation}
DeepSeek R1 was used as the evaluator model. DeepSeek R1, with its deep semantic understanding, rich subject knowledge, and sharp capability in capturing educational intent, is well-suited for evaluating the quality of the generated questions. We embedded the evaluation criteria directly into the evaluation prompts to ensure that the evaluation model could score the generated questions based on clear guidelines and provide detailed evaluation reports. Additionally, to improve the reliability and stability of the evaluation results, we adopted a multi-round voting mechanism to reduce the random error of a single evaluation. Specifically, each sample underwent three independent rounds of scoring, with the mode selected as the final score. In case no mode existed, the arithmetic mean was used as the final score. The evaluation prompt is shown in Figure \ref{fig3}.
\subsection{Experimental Results}
The experimental results are shown in Table ~\ref{all_res}. From the results, it is evident that in the closed-source general-purpose models, Doubao-1.5-thinking-pro exhibited the best overall performance across all dimensions, scoring over 1.9 points in all dimensions except for competence-oriented guidance, where it scored relatively lower. It ranked in the Top 2 for all three subjects. Among open-source general-purpose models, DeepSeek-R1 demonstrated the best overall performance across all subjects. However, Llama-3.1-8B-Instruct performed the worst across all subjects, particularly lagging in the dimensions of question type alignment, question item quality, and solution explanation quality. In the educational models, Spark-X1 from a closed-source setup performed the best overall, though still lagging behind general-purpose models. The open-source educhat-base-002-13b performed the worst, with notably low scores across all dimensions.

\begin{table*}
    \centering
    \footnotesize
    \renewcommand{\arraystretch}{1.3}
    \setlength{\tabcolsep}{3pt}
    \begin{tabular}{c|ccccc|ccccc|ccccc}
    \hline
    \multirow{2}{*}{Model} & 
\multicolumn{5}{c|}{Mathematics} & 
\multicolumn{5}{c|}{Physics} & 
\multicolumn{5}{c}{Chemistry} \\ \cline{2-16}
        & KP & QT & QQ & SQ & CG & KP & QT & QQ & SQ & CG & KP & QT & QQ & SQ & CG \\ \hline
        Qwen3-235B-A22B  & 1.98  & 1.68  & 1.72  & 1.75  & 0.23  & 1.99  & 1.75  & 1.57  & 1.64  & 0.86  & \textbf{2.00}  & 1.72  & 1.64  & 1.69  & 0.71   \\ 
        Qwen3-8B  & 1.99  & 1.80  & 1.81  & 1.79  & 0.22  & 1.97  & 1.91  & 1.74  & 1.58  & 0.71  & 1.99  & 1.89  & 1.68  & 1.51  & 0.54   \\ 
        Qwen3-32B  & 1.98  & 1.95  & 1.87  & 1.84  & 0.23  & \textbf{2.00}  & 1.96  & 1.73  & 1.69  & 0.97  & 1.99  & 1.97  & 1.87  & 1.74  & 0.69   \\ 
        QwQ-32B  & \textbf{2.00}  & 1.90  & 1.89  & 1.87  & 0.21  & \textbf{2.00}  & 1.94  & 1.80  & 1.80  & 0.92  & 1.98  & 1.95  & 1.85  & 1.81  & 0.65   \\ 
        Qwen2.5-72B-Instruct  & 1.95  & 1.80  & 1.84  & 1.74  & 0.18  & 1.97  & 1.87  & 1.72  & 1.57  & 0.65  & 1.96  & 1.92  & 1.79  & 1.63  & 0.60   \\ 
        DeepSeek-V3  & 1.97  & 1.89  & 1.79  & 1.76  & 0.20  & 1.98  & 1.92  & 1.69  & 1.67  & 0.93  & 1.98  & 1.98  & 1.87  & 1.79  & 0.64   \\ 
        DeepSeek-R1  & 1.98  & 1.95  & 1.95  & 1.96  & 0.17  & 1.99  & \textbf{1.99}  & \textbf{1.95}  & 1.94  & 0.77  & 1.99  & 1.93  & 1.91  & 1.94  & 0.73   \\ 
        GLM-Z1-32B  & 1.98  & 1.77  & 1.69  & 1.55  & 0.28  & \textbf{2.00}  & 1.84  & 1.53  & 1.45  & 1.13  & 1.97  & 1.82  & 1.63  & 1.37  & 0.74   \\ 
        GLM-4-32B  & 1.96  & 1.66  & 1.67  & 1.64  & 0.19  & 1.97  & 1.74  & 1.59  & 1.62  & 0.78  & 1.99  & 1.80  & 1.67  & 1.59  & 0.47   \\ 
        GLM-Z1-9B  & 1.98  & 1.21  & 1.57  & 1.53  & 0.23  & 1.94  & 1.37  & 1.25  & 1.10  & 0.76  & 1.94  & 1.16  & 0.94  & 0.94  & 0.53   \\ 
        Yi-34B-Chat  & 1.79  & 1.08  & 1.18  & 0.60  & 0.15  & 1.86  & 1.49  & 1.33  & 0.95  & 0.58  & 1.95  & 1.54  & 1.28  & 0.85  & 0.23   \\ 
        internlm3-8b-instruct  & 1.72  & 0.75  & 1.24  & 1.13  & 0.24  & 1.79  & 1.09  & 1.20  & 1.11  & 0.73  & 1.84  & 1.31  & 1.15  & 1.11  & 0.52   \\ 
        Moonlight-16B-A3B-Instruct  & 1.61  & 1.29  & 1.42  & 1.11  & 0.25  & 1.65  & 1.43  & 1.19  & 0.83  & 0.63  & 1.71  & 1.65  & 1.38  & 0.90  & 0.56   \\ 
        Llama-4-Scout-17B-16E-Instruct  & 1.96  & 1.24  & 1.38  & 1.35  & 0.17  & 1.92  & 1.30  & 0.98  & 1.09  & 0.76  & 1.92  & 0.71  & 0.85  & 0.93  & 0.50   \\ 
        Llama-3.3-70B-Instruct  & 1.78  & 1.78  & 1.46  & 1.19  & 0.19  & 1.83  & 1.68  & 1.41  & 1.11  & 0.77  & 1.91  & 1.71  & 1.39  & 1.14  & 0.51   \\ 
        Llama-3.1-405B-Instruct  & 1.82  & 1.50  & 1.54  & 1.47  & 0.24  & 1.93  & 1.80  & 1.59  & 1.22  & 1.10  & 1.89  & 1.74  & 1.39  & 0.89  & 0.64   \\ 
        Llama-3.1-8B-Instruct  & 1.52  & 0.84  & 0.81  & 0.42  & 0.24  & 1.55  & 0.91  & 0.67  & 0.29  & 0.38  & 1.78  & 0.61  & 0.53  & 0.39  & 0.11   \\ 
        gemma-3-27b-it  & 1.89  & 1.86  & 1.64  & 1.59  & 0.17  & 1.92  & 1.88  & 1.42  & 1.14  & 0.90  & 1.94  & 1.83  & 1.50  & 1.09  & 0.71   \\ 
        gemma-3-12b-it  & 1.87  & 1.23  & 1.39  & 1.39  & 0.23  & 1.86  & 1.54  & 1.24  & 1.00  & 1.06  & 1.91  & 1.78  & 1.30  & 0.81  & 0.73   \\ 
        Mistral-Small-3.1-24B-Instruct  & 1.80  & 1.52  & 1.56  & 1.35  & 0.17  & 1.90  & 1.60  & 1.34  & 1.21  & 0.73  & 1.89  & 1.73  & 1.51  & 1.17  & 0.45   \\ 
        Mistral-Large-Instruct  & 1.86  & 1.63  & 1.68  & 1.57  & 0.17  & 1.90  & 1.81  & 1.69  & 1.42  & 0.57  & 1.97  & 1.89  & 1.75  & 1.49  & 0.32   \\ 
        Phi-4  & 1.85  & 1.86  & 1.61  & 1.49  & 0.14  & 1.91  & 1.88  & 1.58  & 1.26  & 0.63  & 1.91  & 1.88  & 1.60  & 1.22  & 0.37   \\ 
        doubao-1-5-thinking-pro  & 1.99  & 1.96  & 1.96  & 1.97  & 0.25  & \textbf{2.00}  & 1.98  & 1.89  & \textbf{1.96}  & 1.20  & \textbf{2.00}  & 1.93  & 1.91  & \textbf{1.98}  & 0.75   \\ 
        doubao-1.5-vision-pro  & 1.98  & \textbf{1.98}  & 1.90  & 1.90  & 0.18  & 1.98  & 1.98  & 1.79  & 1.78  & 0.81  & 1.99  & 1.95  & 1.89  & 1.88  & 0.61   \\ 
        doubao-1-5-pro-32k  & 1.99  & \textbf{1.98}  & 1.92  & 1.88  & 0.19  & 1.98  & 1.98  & 1.80  & 1.83  & 0.75  & 1.99  & \textbf{2.00}  & 1.93  & 1.91  & 0.59   \\ 
        glm-4-Plus  & 1.89  & 1.69  & 1.65  & 1.56  & 0.19  & 1.83  & 1.77  & 1.56  & 1.31  & 1.18  & 1.91  & 1.93  & 1.57  & 1.17  & 0.77   \\ 
        glm-z1-air  & 1.96  & 1.80  & 1.73  & 1.60  & 0.30  & 1.96  & 1.89  & 1.53  & 1.29  & \textbf{1.31}  & 1.95  & 1.85  & 1.58  & 1.08  & 0.74   \\ 
        qwen-plus-latest  & 1.99  & 1.87  & 1.90  & 1.87  & 0.21  & 1.98  & 1.96  & 1.82  & 1.83  & 0.87  & 1.99  & 1.97  & 1.87  & 1.86  & 0.70   \\ 
        qwen-max-latest  & 1.96  & 1.92  & 1.75  & 1.75  & 0.21  & 1.98  & 1.95  & 1.88  & 1.76  & 0.86  & 1.98  & 1.98  & 1.89  & 1.82  & 0.58   \\ 
        qwq-plus-latest  & 1.98  & 1.76  & 1.79  & 1.80  & 0.23  & \textbf{2.00}  & 1.67  & 1.57  & 1.73  & 0.85  & \textbf{2.00}  & 1.74  & 1.60  & 1.67  & 0.67   \\
        o4-mini  & 1.98  & 1.94  & \textbf{1.97}  & 1.94  & 0.19  & 1.96  & 1.94  & 1.80  & 1.76  & 0.81  & 1.97  & 1.92  & 1.84  & 1.73  & 0.62   \\
        o3-mini  & 1.98  & 1.93  & 1.97  & \textbf{1.98}  & 0.15  & 1.99  & 1.95  & 1.89  & 1.89  & 0.79  & \textbf{2.00}  & 1.95  & 1.87  & 1.82  & 0.71   \\
        gpt-4.1  & 1.94  & 1.94  & 1.71  & 1.69  & 0.18  & 1.98  & 1.95  & 1.83  & 1.76  & 0.88  & 1.99  & 1.96  & 1.81  & 1.73  & 0.47   \\ 
        GPT-4o  & 1.96  & 1.82  & 1.76  & 1.72  & 0.21  & 1.99  & 1.90  & 1.80  & 1.75  & 0.85  & 1.99  & 1.95  & 1.80  & 1.69  & 0.49   \\ 
        Claude 3.7 Sonnet  & 1.87  & 1.73  & 1.65  & 1.52  & 0.19  & 1.93  & 1.88  & 1.46  & 1.29  & 1.16  & 1.97  & 1.96  & 1.52  & 1.07  & 0.83   \\ 
        Claude 3.5 Haiku  & 1.86  & 1.74  & 1.51  & 1.27  & 0.21  & 1.85  & 1.81  & 1.42  & 0.96  & 1.17  & 1.87  & 1.93  & 1.42  & 0.87  & 0.91   \\
        Gemini 2.5 Pro Preview  & 1.98  & 1.93  & 1.92  & 1.93  & 0.16  & 1.98  & 1.95  & 1.91  & 1.90  & 1.00  & 1.99  & 1.95  & \textbf{1.94}  & 1.90  & \textbf{0.92}   \\
        Gemini 2.5 Flash Preview  & \textbf{2.00}  & 1.95  & 1.96  & 1.96  & 0.22  & 1.99  & 1.96  & 1.93  & 1.94  & 0.87  & \textbf{2.00}  & 1.96  & 1.92  & 1.97  & 0.67   \\
        Moonshot-v1-32K  & 1.87  & 1.43  & 1.45  & 1.37  & 0.31  & 1.91  & 1.71  & 1.49  & 1.18  & 1.01  & 1.95  & 1.90  & 1.56  & 1.33  & 0.72   \\
        Hunyuan-Large  & 1.94  & 1.91  & 1.88  & 1.78  & 0.23  & 1.92  & 1.90  & 1.67  & 1.55  & 0.81  & 1.93  & 1.81  & 1.80  & 1.68  & 0.57   \\
        Yi-Lightning  & 1.91  & 1.89  & 1.75  & 1.57  & 0.14  & 1.98  & 1.95  & 1.71  & 1.67  & 0.95  & 1.97  & 1.96  & 1.80  & 1.65  & 0.70   \\ 
        educhat-base-002-13b  & 1.09  & 0.68  & 0.75  & 0.45  & 0.18  & 1.25  & 0.81  & 0.63  & 0.22  & 0.19  & 1.30  & 0.73  & 0.55  & 0.16  & 0.13   \\ 
        educhat-sft-002-13b  & 1.89  & 1.64  & 1.45  & 0.58  & \textbf{0.42}  & 1.94  & 1.63  & 1.03  & 0.25  & 0.66  & 1.88  & 1.58  & 0.85  & 0.16  & 0.42   \\ 
        Confucius-o1  & 1.92  & 1.58  & 1.74  & 1.60  & 0.25  & 1.96  & 1.61  & 1.47  & 1.32  & 0.87  & 1.95  & 1.80  & 1.67  & 1.45  & 0.78   \\ 
        Spark-X1  & 1.99  & 1.63  & 1.70  & 1.76  & 0.22  & \textbf{2.00}  & 1.71  & 1.50  & 1.56  & 1.13  & 1.98  & 1.67  & 1.57  & 1.64  & 0.74   \\ 
        Spark-lite  & 1.39  & 1.21  & 1.04  & 0.45  & 0.13  & 1.61  & 1.43  & 1.11  & 0.45  & 0.39  & 1.72  & 1.66  & 0.96  & 0.33  & 0.35   \\ \hline
    \end{tabular}
    \caption{Model scores are evaluated using DeepSeek-R1. The maximum score in each dimension is highlighted in bold, and the full names of the evaluation dimensions are provided in Section 4.1.}
    \label{all_res}
\end{table*}
\subsubsection{Dimension Analysis}
Models showed overall stability in their performance on fundamental understanding tasks, with minimal differences between subjects. For knowledge point alignment and question type alignment, both closed-source and open-source models generally scored highly. For example, in top-performing general-purpose models such as o4-mini, QwQ-32B, and Doubao-1.5 series, the scores for all three subjects were close to full marks. In educational models, these two dimensions also generally outperformed the other three. This indicates that mainstream LLMs have a strong ability to recognize and map the basic structure of questions and their core assessment points.

In tasks requiring higher reasoning and logical ability, closed-source models and larger parameter open-source models showed an edge. In the question item quality and solution explanation quality dimensions, there was clear performance stratification, with models like Doubao-1.5-thinking-pro and DeepSeek-R1 achieving scores above 1.9, while the worst-performing models scored no higher than 0.8. Meanwhile, in educational models, the highest-scoring models still fell short when compared to the top general-purpose models. A significant trend observed was that models performed better in mathematics due to the abundant training data available in this domain, often scoring higher in mathematics than in physics and chemistry. This highlights the reasoning advantage of general-purpose models, while educational models, despite focusing more on educational scenarios, still struggle with complex cognitive tasks.

The competence-oriented guidance dimension was the weakest for all models. Scores for mathematics were particularly low, while physics and chemistry performed relatively better. This suggests that models still lack a strong ability to understand the educational intent behind question design, especially in mathematics. Nearly no model scored above 0.3 in mathematics, while the best-performing models in physics and chemistry reached 0.9. Most models scored around 0.7 in these dimensions, reflecting the current gap in models' ability to generate contextually relevant questions, especially in more abstract subjects like mathematics.

\subsubsection{Subject-wise Analysis}
\paragraph{Mathematics}
The top three models in terms of overall performance were Doubao-1.5-thinking-pro, Gemini-2.5-flash-preview, and o4-mini, with total scores exceeding 8. The bottom three models were educhat-base-002-13b, Llama-3.1-8B-Instruct, and Spark-Lite, with scores of 3.14, 3.83, and 4.22, respectively. QwQ-32B scored full marks for knowledge point alignment, and Doubao-1.5-pro-32k and Doubao-1.5-vision-pro scored highest for question type alignment, while DeepSeek-R1 achieved the highest score of 1.97 for question item quality.

\paragraph{Physics}
The top three models were Doubao-1.5-thinking-pro, Gemini-2.5-pro-preview, and Gemini-2.5-flash-preview, with scores of 9.02, 8.74, and 8.69, respectively. Doubao-1.5-thinking-pro achieved full marks for knowledge point alignment, and DeepSeek-R1 ranked first in both question type alignment and question item quality, with scores of 1.99 and 1.95, respectively. The top score for competence-oriented guidance was 1.31 by glm-z1-air.

\paragraph{Chemistry}
The top three models were Gemini-2.5-pro-preview, Doubao-1.5-thinking-pro, and Gemini-2.5-flash-preview, with scores of 8.71, 8.58, and 8.51, respectively. Doubao-1.5-thinking-pro, o3-mini, and Qwen3-235B-A22B achieved full marks for knowledge point alignment, and Doubao-1.5-pro-32k scored the highest for question type alignment. Gemini-2.5-pro-preview scored the highest for question item quality at 1.94.

While the top models from major companies perform similarly across subjects, the variations are mostly in the competence-oriented guidance dimension. This suggests that while there are no significant differences in the overall performance of top models across middle school subjects, the ability to generate contextually relevant and application-driven questions remains a critical area for improvement.

\subsection{Human Study}
Given the inherent uncertainty and randomness associated with LLMs as evaluators, we introduced an expert manual evaluation to verify the validity and reliability of the automated evaluation algorithm. We randomly selected 100 test samples from the mathematics evaluation set and obtained results from GLM-Z1-9B, Spark-X1, and o4-mini. Six experienced middle school mathematics teachers were invited to perform independent evaluations, adhering to the same dimensions and standards as the automated evaluation.

To assess the consistency between human evaluators and the automated evaluator, we employed two different evaluation methodologies. First, we conducted a Score-level Consistency assessment (SC) by directly computing the numerical differences between human evaluator scores and automated evaluator score. Second, we performed a Ranking-level Consistency assessment (RC) by analyzing the correlation between human evaluators' rankings and the automated evaluator's rankings of the three models, using Spearman's rank correlation coefficient. SC and RC can be formulated as:
\begin{equation}
\begin{split}
\label{eq:combined}
  &SC{_j} = 1 -\frac{1}{K} \sum_{k=1}^{K}
  \\& \left( \frac{1}{N} \sum_{i=1}^{N} \left| S^{model_k}(Q_i, D_j) - S^{human_k}(Q_i, D_j) \right| \right)
\end{split}
\end{equation}
\begin{equation}
  \label{eq:example}
  RC_{j}=1-\frac{6\sum_{K=1}^{K}{{{d}_{K}^{2}}}}{K\left ({{{K}^{2}}-1}\right )}
\end{equation}
where ${{S}^{huma{{n}_{k}}}}\left ({{{Q}_{i}},{{D}_{j}}}\right )$ represents the scores given by the human evaluator for the $k$-th model on $i$-th question $Q$ with $j$-th dimension $D$ and $N=100$ is the total number of samples.$K=3$ is the number of models.$d$ represents the difference in ranks for a given model on a specific dimension.
\begin{table}
    \centering
    \small
    \setlength{\tabcolsep}{5.3pt}
    \renewcommand{\arraystretch}{1.4}
    \begin{tabular}{cccccc} 
    \hline
     & KP & QT & QQ & SQ & CG \\ \hline
    SC & 0.9650 & 0.9733 & 0.8983 & 0.8850 & 0.9197 \\ 
    RC & 1.0000 & 1.0000 & 1.0000 & 0.5000 & 1.0000 \\ \hline
    \end{tabular}
    \caption{Scores for SC and RC across different dimensions are obtained by calculating LLM and human scores.A higher SC indicates a stronger agreement between the model's and human's scores, while higher RC values signify better alignment between the two ranking systems.}
    \label{sc and srcc}
\end{table}

The Score Consistency results showed that DeepSeek-R1 achieved over 88\% consistency with human scores across all dimensions, with the highest consistency in question type alignment, reaching 97\%. The $SRCC$ results showed a perfect correlation $1$ for most dimensions, with the solution explanation quality dimension having a correlation of 0.5. This demonstrates that our automated evaluation framework aligns closely with expert human evaluation, confirming the effectiveness and reliability of the proposed framework.

\section{Conclusion}
This paper introduces EQGBench, a benchmark for evaluating the educational question generation capabilities of Large Language Models (LLMs). EQGBench features a high-quality dataset of 900 structured instructions across mathematics, physics, and chemistry, reflecting diverse, real-world user needs. Its core is a multi-dimensional framework, aligned with pedagogical goals, that assesses models on five key metrics. The framework's automated pipeline demonstrates high reliability and consistency, as validated against expert teacher assessments.

Our evaluation of 46 mainstream LLMs reveals that while leading models possess strong foundational capabilities, they struggle to generate questions with deep pedagogical intent. We believe EQGBench is a valuable resource for the academic community that will guide the future optimization of LLMs for education.

\bibliography{main}

\end{document}